%% file: bmvc_review.tex
\documentclass{bmvc2k}

\title{RGFVR: Reference-Guided Face Video Restoration with Flow Matching}

\addauthor{Cem Eteke\textsuperscript{*}}{cem.eteke@tum.de}{1}
\addauthor{Batuhan Tosun\textsuperscript{*}}{batuhan.tosun@tum.de}{1}
\addauthor{Eckehard Steinbach}{eckehard.steinbach@tum.de}{1}

\addinstitution{
  Chair of Media Technology\\
  Munich Institute of Robotics and Machine Intelligence\\
  School of Computation, Information, and Technology\\
  Technical University of Munich\\
  80333 Munich, Germany\\[4pt]
  \scriptsize{\textsuperscript{*}Equal contribution}
}

\runninghead{Eteke, Tosun}{RGFVR}

\def\ie{\emph{i.e}\bmvaOneDot}

\usepackage{graphicx}
\usepackage{booktabs}
\usepackage{hyperref}
\usepackage{multirow}
\usepackage{bbding,amsmath,amssymb,graphicx,amsfonts,pifont,mathtools,booktabs,diagbox,tikz,xcolor,bm,colortbl,enumitem}

\newcommand{\tightbox}[2]{%
  {\setlength{\fboxsep}{0.25pt}\colorbox{#1}{#2}}%
}

\definecolor{artifact}{HTML}{FF887B}
\definecolor{zoom}{HTML}{B4E3FF}
\definecolor{Gold}{RGB}{255,215,0}
\definecolor{Silver}{RGB}{192,192,192}
\definecolor{Bronze}{RGB}{205,127,50}
\definecolor{red}{RGB}{255,0,0}
\colorlet{GoldSoft}{Gold!25}
\colorlet{SilverSoft}{Silver!25}
\colorlet{BronzeSoft}{Bronze!25}
\newcommand{\best}[1]{\tightbox{GoldSoft}{\textbf{#1}}}
\newcommand{\second}[1]{\tightbox{SilverSoft}{#1}}
\newcommand{\third}[1]{\tightbox{BronzeSoft}{#1}}

\definecolor{red}{RGB}{255,0,0}

\definecolor{blue}{RGB}{0,0,255}

\definecolor{green}{RGB}{0,255,0}

\begin{document}

\maketitle

\begin{figure*}[h!]
\begin{center}
\includegraphics[width=\textwidth]{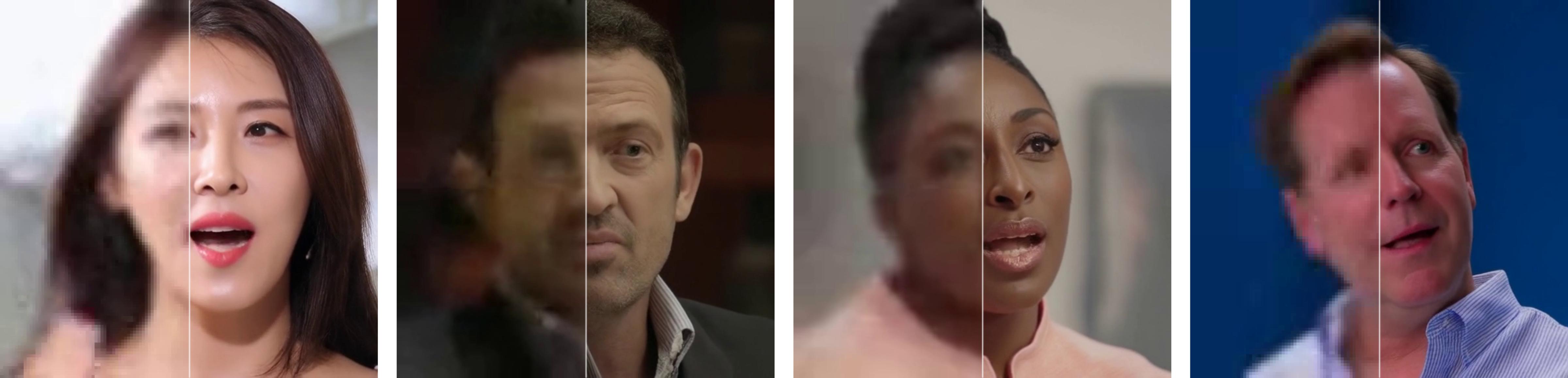}
\end{center}
   \caption{Loss of facial details under degradations and restoration using our method.}
\label{fig:teaser}
\end{figure*}

\vspace{-10pt}

\begin{abstract}
    Face video restoration from degraded observations is challenging, as it requires simultaneously recovering visual fidelity, temporal consistency, and subject identity. Existing approaches are often either reference-free, which can lead to identity loss when person-specific facial details are lost, or subject-specific, which limits generalization to unseen identities. We propose a subject-agnostic, reference-guided framework for identity-preserving face video restoration. Our method introduces bimodal perceptual-descriptive identity conditioning into a pretrained flow-based text-to-video generator and employs a two-stage training strategy to strengthen identity guidance during restoration. Experiments show that our approach improves restoration fidelity, temporal consistency, and identity preservation, achieving superior performance under challenging video degradations, including downsampling, blur, noise, and compression artifacts. The code is available under: \url{https://github.com/batuhanntosun/RG-FVR}.
\end{abstract}

\section{Introduction}
\label{sec:intro}

Restoring high-quality face videos from severely degraded observations is a challenging inverse problem with broad applications in human-centered visual computing. Unlike generic video restoration, face video restoration additionally requires preserving the subject’s identity, including characteristic facial structure, feature proportions, and person-specific appearance. This requirement becomes especially challenging under severe degradations such as low resolution, blur, compression artifacts, noise, and combinations thereof, where identity cues may be heavily obscured. Fig.~\ref{fig:teaser} illustrates examples of such identity-obscuring degradations. In these regimes, visually plausible restoration does not necessarily imply identity preservation. Consequently, generic restoration methods often fail to faithfully recover identity.

To better constrain restoration under severe degradation, face image restoration methods incorporate generative priors~\cite{gpen, glean, gfpgan, pulse}, codebook-based priors~\cite{codeformer, vqfr}, and, more recently, diffusion priors~\cite{difface, diffbir, osdface} to enable realistic facial reconstruction. However, these approaches operate primarily in the image domain and therefore do not address the temporal consistency required in video restoration. Recent face video restoration methods extend image-based restoration to the temporal setting~\cite{keep0, pgtformer, svfr}. However, most remain reference-free, relying solely on degraded observations to infer and restore identity. Under severe degradation, this often leads to low identity fidelity and temporal inconsistency. A natural way to alleviate this ambiguity is to condition restoration on a high-quality reference image of the target subject. However, existing reference-guided approaches are predominantly image-based~\cite{restorerid, refldm, instantrestore}, while the few video-domain methods typically require subject-specific optimization, limiting scalability to unseen identities~\cite{rivalmethod}. Furthermore, identity conditioning in pretrained diffusion or flow-based generators is often overshadowed by the strong generative prior of the backbone, due to overparameterization, thereby weakening its influence during restoration.

To fill this gap, we propose a subject-agnostic reference-guided face video restoration methodology. Our method introduces bimodal perceptual-descriptive identity conditioning for a pretrained flow-based video generator, enabling robust identity guidance under severe degradations while maintaining temporal coherence. To mitigate the pretrained backbone's overshadowing at the expense of identity fidelity, we introduce a two-stage training strategy that strengthens identity conditioning during optimization. Unlike prior subject-specific approaches, our method generalizes to unseen identities without requiring per-subject optimization. Extensive experiments demonstrate that the proposed approach improves pixel-level, perceptual, and identity fidelity, achieving state-of-the-art performance.

To summarize, in this work, we contribute:

\begin{itemize}
    \item A subject-agnostic reference-guided method for identity-preserving face video restoration.
    \item Bimodal perceptual-descriptive identity conditioning for reference guidance.
    \item A two-stage training strategy that strengthens identity preservation while maintaining restoration fidelity and temporal consistency.
\end{itemize}

\section{Related work}
\label{sec:relatedwork}

Video restoration aims to recover high-quality temporally consistent frames from degraded inputs. Early approaches relied on explicit temporal modeling to exploit inter-frame redundancy~\cite{edvr,basicvsr,vrt}. More recently, restoration methods have increasingly adopted pretrained generative backbones, particularly diffusion models, due to their strong natural-image priors and perceptual synthesis capabilities~\cite{diffusionbeatsgan}. Early diffusion-based methods adapted pretrained image restoration models to videos through temporal conditioning~\cite{stablevsr,upscaleavideo}, while more recent approaches directly build on video diffusion models that model temporality~\cite{seedvr0}. Despite strong perceptual quality, these methods do not explicitly enforce the structural and identity constraints required for face video restoration.

A central challenge in face restoration is preserving facial structure and attributes, \ie, the identity. Early methods relied on structural priors such as landmarks, parsing maps, and 3D face geometry~\cite{fsrnet,psfrgan,facial3dfsr,face3dprior}. GAN-based approaches later leveraged pretrained face generators either through latent optimization~\cite{pulse} or feed-forward prior injection~\cite{gpen,gfpgan,sgpn}. More recent methods instead adopted codebook-based discrete facial priors~\cite{codeformer,vqfr}, which improve robustness by constraining restoration to a learned dictionary of facial representations. However, the codebook's finite capacity limits generalization under out-of-distribution degradations. Recent work has therefore increasingly shifted toward pretrained diffusion models, which provide stronger generative priors and improved perceptual synthesis capabilities. DifFace restored degraded faces by mapping them to intermediate diffusion timesteps~\cite{difface}, while DiffBIR combined coarse restoration with controllable diffusion refinement~\cite{diffbir}. Compared with face image restoration, face video restoration remains relatively underexplored. KEEP extended codebook-based restoration with Kalman-based temporal consistency~\cite{keep0}, PGTFormer introduced semantic parsing guidance for spatiotemporal transformers~\cite{pgtformer}, and SVFR extended video diffusion transformers with refinement and landmark regularization for temporal consistency~\cite{svfr}. However, these methods are reference-free and therefore struggle to reliably recover identity under severe degradation. This limitation motivates reference-guided restoration.

Reference-guided identity-preserving generation conditions restoration on facial attributes extracted from a high-quality reference image. Existing approaches can broadly be divided into subject-agnostic methods that generalize across identities without per-subject optimization~\cite{instantid,idanimator,consisid}, and personalized methods that adapt the model to a specific subject~\cite{dreambooth}. Early image-based approaches relied on explicit correspondence through face warping, landmark alignment, or dictionary matching~\cite{gfrnet,asffnet,dmdnet}. More recent diffusion-based methods instead condition directly on reference features~\cite{restorerid,instantrestore,refldm}, while personalized methods such as PFStorer~\cite{pfstorer} optimize restoration for a single target identity. As with reference-free methods, reference-guided face video restoration is underexplored, and existing methods remain limited to subject-specific approaches. For example, Show \& Polish~\cite{rivalmethod} extends diffusion-based video restoration through subject-specific low-rank finetuning. In contrast, we propose, to the best of our knowledge, the first subject-agnostic framework for identity-preserving face video restoration. Our method introduces bimodal perceptual-descriptive identity conditioning to a pretrained flow-based video generator. Furthermore, unlike prior reference-guided approaches that rely primarily on architectural conditioning alone, we introduce a two-stage training strategy that strengthens identity conditioning against the strong generative prior of the pretrained backbone.

\section{Methodology}
\label{sec:methodology}

\begin{figure*}[t!]
\begin{center}
 \includegraphics[width=1.0\textwidth]{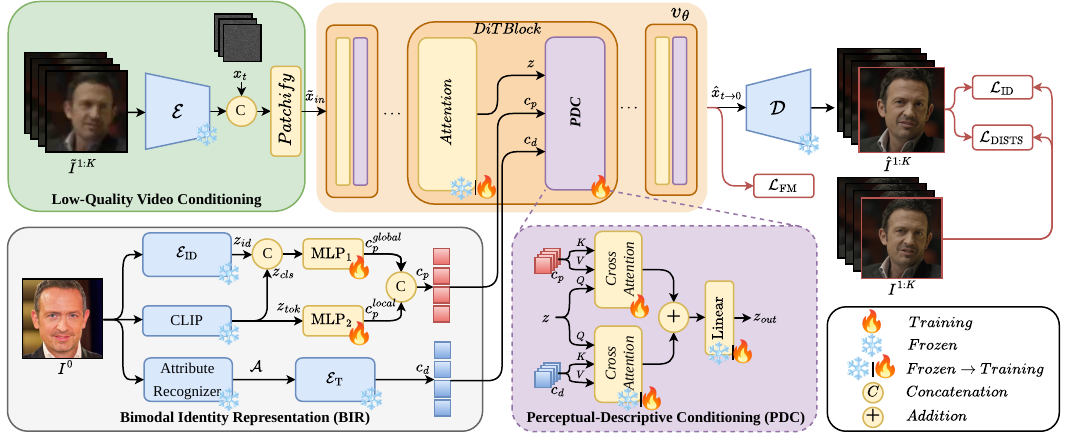}
\end{center}
   \caption{Overview of our two-stage reference-guided identity-preserving face video restoration framework. We represent identity from the reference image $I^0$ using BIR: a face encoder $\mathcal{E}_\mathrm{ID}$ and CLIP provide local-global decomposed perceptual features $c_p$, while a face descriptor and a text encoder $\mathcal{E}_\mathrm{T}$ produce descriptive features $c_d$. In Stage 1, we condition the DiT on identity via our PDC module, optimizing only the conditioning modules for the identity loss $\mathcal{L}_\mathrm{ID}$ and the flow-matching loss $\mathcal{L}_\mathrm{FM}$. In Stage 2, we enable restoration. We use the Encoder $\mathcal{E}$ on degraded frames $\tilde{I}^{1:K}$ to obtain the degraded latents $\tilde{x}$. We channel-concatenate $\tilde{x}$ with the noisy latent $\tilde{x}_t$, forming the input $\tilde{x}_{in}$, optimizing the perceptual loss $\mathcal{L}_\mathrm{DISTS}$.}
\label{fig:pipeline}
\end{figure*}

Our proposed method leverages the generative prior of a pre-trained video Diffusion Transformer (DiT) backbone~\cite{wan0}. We derive perceptual-descriptive identity representations from a high-quality reference frame using the Bimodal Identity Representation (BIR), which decouples \textit{global} and \textit{local} features. We then inject these representations into the DiT blocks via Perceptual-Descriptive Conditioning (PDC), enabling joint modeling. We train the model in two stages: first, identity-conditioned video synthesis, and then identity-preserving restoration. We illustrate the overall framework in Fig.~\ref{fig:pipeline}.

\subsection{Preliminaries}

\subsubsection{Rectified flow}
Rectified flow is a continuous-time generative framework that learns a velocity field transporting noise to data along straight trajectories~\cite{liu2022flow, esser2024flow, lipman2022flow}. Given a clean sample $x_0$ and noise $x_1~\sim~\mathcal{N}(0, I)$, the intermediate state at time $t \in [0,1]$ is
\begin{equation}
x_t = (1-t)x_0 + t x_1 .
\label{eq:rectified_flow}
\end{equation}
A model $v_\theta$ is trained to predict the transport direction with
\begin{equation}
\mathcal{L}_{\mathrm{FM}} =
\mathbb{E}_{t,x_0,x_1}
\left[
\|v_\theta(x_t,c,t) - (x_1-x_0)\|^2
\right],
\label{eq:flow_matching}
\end{equation}
where $c$ denotes optional conditioning information, e.g., text, class labels, or reference features. Generation starts from $x_1 \sim \mathcal{N}(0,I)$ and integrates
\begin{equation} \label{eq:ode}
\frac{d x_t}{dt}=v_\theta(x_t,c,t)
\end{equation}
from $t=1$ to $t=0$ to obtain a sample $x_0$ consistent with $c$. Although rectified flow differs from diffusion models~\cite{ddpm} and latent diffusion models (LDMs)~\cite{ldm} in formulation, these approaches share iterative generation processes based on pretrained generative priors. Since our method builds on a pretrained flow-based video generator closely related to diffusion architectures, we occasionally refer to these models jointly as diffusion or flow-based generators.

\subsubsection{Diffusion transformer} \label{sec:dit}

We use a pretrained DiT as the velocity model $v_\theta$ for scalable video generation, thereby enabling a strong natural image prior~\cite{peebles2023scalable}. Here we utilize the LDM formulation, where given an input video $I^{1:K} \in \mathbb{R}^{K \times H \times W \times 3}$, an encoder $\mathcal{E}$ maps the frames to downsampled latents $x_0=\mathcal{E}(I^{1:K}) \in \mathbb{R}^{K' \times H' \times W' \times D}$, and noisy latents $x_t$ are obtained as in Eq.~\ref{eq:rectified_flow}. The DiT processes $x_t$ first via \textit{patchification}, which divides the latents into tokens and extracts features via spatiotemporal convolution, resulting in spatiotemporal token $x_{in} \in \mathbb{R}^{L \times D}$. We present $L$ for simplicity, which comprises the spatial and temporal tokens. The DiT backbone passes $x_{in}$ through spatial and temporal self-attention blocks, while injecting conditioning signals $c$ via cross-attention. For hidden tokens $z \in \mathbb{R}^{L \times D_h}$ of a DiT block and conditioning tokens $c \in \mathbb{R}^{M \times D_c}$, cross-attention is
\begin{equation} \label{eq:cross_attention}
Attention(Q,K,V)=softmax\left(\frac{QK^T}{\sqrt{d}}\right)V,
\end{equation}
where
\begin{equation}
Q=zW^Q,\quad K=cW^K,\quad V=cW^V,
\end{equation}
with
\begin{equation}
W^Q \in \mathbb{R}^{D_h \times d},\quad
W^K \in \mathbb{R}^{D_c \times d},\quad
W^V \in \mathbb{R}^{D_c \times d}.
\end{equation}

This allows each latent token to adaptively retrieve relevant conditioning information. In our case, we utilize a pretrained text-to-video DiT~\cite{wan0}, hence $c$ comprises text embeddings. 

Finally, the DiT predicts the velocity and is trained with Eq.~\ref{eq:flow_matching}. During inference, the latent trajectory is iteratively integrated from $x_1$ to $\hat{x}_0$ to solve Eq.~\ref{eq:ode}, and a decoder $\mathcal{D}$ reconstructs the generated video

\begin{equation}
\hat{I}^{1:K}=\mathcal{D}(\hat{x}_0).
\end{equation}

\subsection{Degraded video conditioning} \label{sec:deg_cond}

We aim to reconstruct high-quality frames from a degraded video input. As presented in Fig.~\ref{fig:pipeline}, we first encode the degraded video $\tilde{I}^{1:K}$ into a latent representation $\tilde{x}$. These features are channel-concatenated with the noisy latent $x_t$ at diffusion timestep $t$ and projected via an input layer that extends the first convolutional layer of patchification via
\begin{equation} \label{eq:input_proj}
    \tilde{x}_{in} = conv([x_t~~\tilde{x}]; [W^{in}~~W^{in'}], [b^{in}~~b^{in'}])
\end{equation}
where $W^{in}$ and $b^{in}$ denote the pretrained convolutional kernels and biases, while $W^{in'}$ and $b^{in'}$ are the newly introduced parameters for degraded-video features, initialized as zeros. We extend only the first spatiotemporal convolutional layer as shown, and $\tilde{x}_{in}$ is the result of the entire patchification. This gives the network access to the spatial structure and content of the degraded input without interfering with the initial model behavior. While degraded-video conditioning preserves spatial structure and motion, we motivated above that the degradations might lead to identity loss.

\subsection{Bimodal identity representation} \label{sec:bir}

Identity preservation requires capturing facial characteristics across multiple modalities. One modality is a high-level symbolic language description, which aligns naturally with a pretrained text-to-video DiT. However, such descriptions are propositional: terms such as "brown eyes" specify semantic attributes, but not fine-grained perceptual details such as the exact shade or texture. Hence, identity also benefits from complementary perceptual representations. To capture identity at both levels, we present Bimodal Identity Representation (BIR). Given a high-quality reference frame $I^0$, BIR extracts descriptive, namely $c_d$, and perceptual local-global, namely $c_p$, identity features.

\subsubsection{Descriptive identity}
Text-based descriptive identity conditioning enables adapting a pretrained text-to-video model and was shown to improve the identity fidelity~\cite{rivalmethod}. To that end, we extract the descriptive features $c_d$ from reference $I^0$ using a facial attribute recognition model, which outputs a set of likelihoods $p(a_i~|~I_{ref})$ for each attribute $a_i$~\cite{farl}. We then formulate a subject-specific attribute set $\mathcal{A}$ with a threshold $\tau$,

\begin{equation}
\mathcal{A} = \{\, a_i \mid p(a_i~|~I_{ref}) > \tau \,\} \text{.}
\end{equation}

We convert this set $\mathcal{A}$ to a language symbol-based description of the identity by using the word that corresponds to each attribute. We utilize the text encoder $\mathcal{E}_\mathrm{T}$ of the DiT backbone to extract the descriptive conditioning 

\begin{equation} \label{eq:cd}
    c_d = \mathcal{E}_\mathrm{T}(\mathcal{A}),
\end{equation} both the attribute model $p(.|I_{ref})$ and the $\mathcal{E}_\mathrm{T}$ always remain frozen.

\subsubsection{Perceptual identity}

Perceptual identity is captured through both facial and visual representations~\cite{consisid, pulid}. In this context we extract from the reference frame $I^0$ the face representations using a face encoder $\mathcal{E}_\mathrm{ID}:\mathbb{R}^{H \times W \times 3} \rightarrow \mathbb{R}^{D_{id}}$ 

\begin{equation} \label{eq:face_id}
    z_{id} = \mathcal{E}_\mathrm{ID}(I^0).
\end{equation}

Furthermore, we utilize the \texttt{[CLS]} token $z_{cls} \in \mathbb{R}^{D_{cls}}$ of CLIP as the visual representation~\cite{clip0}. We combine these two representations into a \textit{global} perceptual identity via a learned function $\mathrm{MLP}_1: \mathbb{R}^{D_{id}+D_{cls}} \rightarrow \mathbb{R}^{D_c}$

\begin{equation}\label{eq:mlp1}
    c_p^{global} = \mathrm{MLP}_1\!\left(
        \begin{bmatrix}
            z_{id}~~z_{cls}
        \end{bmatrix}
    \right).
\end{equation}

Even though $c_p^{global}$ encodes the global visual features, it lacks the spatial granularity for high-fidelity restoration. We therefore extract the tokens from a late stage transformer block of the CLIP vision encoder $z_{tok} \in \mathbb{R}^{M_{tok} \times D_{tok}}$ and project the token features via a second learnable function $\text{MLP}_2: \mathbb{R}^{D_{tok}} \rightarrow \mathbb{R}^{D_c}$, forming the \textit{local} representations

\begin{equation}\label{eq:mlp2}
    c_p^{local} = \mathrm{MLP}_2(z_{tok})\text{.}
\end{equation}

We finally create the complete local-global perceptual tokens $c_p \in \mathbb{R}^{(M_{tok} + 1) \times D_c}$ by concatenating the representations along the sequence dimension

\begin{equation} \label{eq:cp}
c_p =
    \begin{bmatrix}
        c_p^{global}~;~c_p^{local}
    \end{bmatrix}.
\end{equation}

As a result, we extract the descriptive identity features $c_d$ (Eq.~\ref{eq:cd}) as well as local-global perceptual features $c_p$ (Eq.~\ref{eq:cp}). Together, the descriptive and perceptual representations provide complementary identity features, enabling robust identity conditioning under severe degradations. We present these steps in Fig.~\ref{fig:pipeline} inside the Bimodal Identity Representation block.

\subsection{Perceptual-descriptive conditioning} \label{sec:pdc}

Descriptive conditioning is enabled through text conditioning of the DiT backbone. The additional perceptual modality requires extending the pretrained text-to-video DiT. To condition with the bimodal identity representation, we present Perceptual-Descriptive Conditioning (PDC). PDC enables conditioning on the decoupled perceptual-descriptive identity representations via parallel cross-attention at each DiT block~\cite{ipadapter}. We display the PDC steps in Fig.~\ref{fig:pipeline}.

Concretely, we extend the text-only cross-attention formulation of the backbone for perceptual identity conditioning. We compute the query from the hidden feature tokens $z$ produced by the preceding self-attention block. We use the cross-attention formulation shown in Eq.~\ref{eq:cross_attention} to condition on bimodal identity features. Projections of the local-global perceptual features $c_p$ and the descriptive features $c_d$ are used to extract the key and the value features as
\begin{equation} \label{eq:per_cond}
K_d = c_d W^K_d + b_d^k, \quad V_d = c_d W^V_d + b_d^V, \qquad K_p = c_p W^K_p + b_p^K, \quad V_p = c_p W^V_p + b_p^V.
\end{equation}
The weights $W^Q$, $W^K_d$, $W^V_d$, and biases $b_d^K$, $b_d^V$ are inherited from the pre-trained backbone, while $W_p^K \in \mathbb{R}^{D_c \times d}$, $W_p^V \in \mathbb{R} ^ {D_c \times d}$, $b_p^V \in \mathbb{R} ^ d$, and $b_p^V \in \mathbb{R} ^ d$ are introduced. We further utilize normalization layers from the backbone after the projections~\cite{wan0}. Parallel cross-attention is then computed over perceptual-descriptive conditioning:
\begin{equation}
z_d = Attention(Q, K_d, V_d), \qquad z_p = Attention(Q, K_p, V_p).
\end{equation}
We finally combine the representations to be passed into the next DiT block after projecting with the output projection:
\begin{equation} \label{eq:z_new}
z \leftarrow(z_d + z_p) W^O + b^O,
\end{equation} where weights $W^O \in \mathbb{R}^{d \times D_h}$ and biases $b^O\in \mathbb{R}^{D_h}$ are inherited from the pretrained backbone. We display these steps in Fig.~\ref{fig:pipeline}.

\subsection{Two-stage training and objective functions} \label{sec:training_stages}

Jointly fine-tuning the DiT backbone for both video restoration and identity preservation yields reduced identity fidelity, as the large-scale pretrained DiT backbone overshadows the identity conditioning. To address this issue, we propose a two-stage training approach where the first stage initializes the introduced cross-attention weights for identity conditioning and the second stage finetunes the entire model for restoration. This ordering first lets the introduced identity-conditioning layers learn an identity signal before the restoration objective updates the full pretrained backbone. We displayed this distinction in Fig.~\ref{fig:pipeline} where $Training$ denotes Stage 1 and $Frozen \rightarrow Training$ denotes Stage 2.

\subsubsection{Stage 1: identity conditioning} ~\label{sec:stage_1}

In Stage 1, we train only the perceptual identity-conditioning parameters: $\mathrm{MLP}_1$ in Eq.~\ref{eq:mlp1}, $\mathrm{MLP}_2$ in Eq.~\ref{eq:mlp2}, and the perceptual key-value projections $W_p^K,W_p^V$ in Eq.~\ref{eq:per_cond}. Descriptive identity-conditioning parameters remain frozen, as they are inherited from the pretrained backbone.

At this stage, the model is trained as an identity-conditioned generator without restoration supervision, providing a strong initialization for the newly introduced modules. To enforce identity fidelity, in addition to the flow matching loss in Eq.~\ref{eq:flow_matching}, we use an identity loss. In order to compute this identity loss, we first approximate the clean latent using a first-order estimation

\begin{equation}
    \hat{x}_{t\rightarrow 0}=x_t-t\,v_\theta(x_t,c_d,c_p,t),
\end{equation} which is decoded as $\hat{I}^{1:K}=\mathcal{D}(\hat{x}_{t\rightarrow 0})$, and compute the average cosine distance between reconstructed and ground-truth identity features extracted as in Eq.~\ref{eq:face_id}:
\begin{equation} \label{eq:id_loss}
    \mathcal{L}_\mathrm{ID}= \mathbb{E}_{t,x_0,x_1}
    \left[
    1-
    \frac{\mathcal{E}_\mathrm{ID}(I^{1:K})\cdot \mathcal{E}_\mathrm{ID}(\hat{I}^{1:K})}
    {\|\mathcal{E}_\mathrm{ID}(I^{1:K})\|\,\|\mathcal{E}_\mathrm{ID}(\hat{I}^{1:K})\|}
\right].
\end{equation}

Since identity estimates become unreliable at high noise levels, we use the common practice of weighting the additional loss~\cite{refldm}. As a result, the Stage 1 objective becomes
\begin{equation} \label{eq:stage_1_loss}
\mathcal{L}_\mathrm{Stage\text{-}1}
=
\mathcal{L}_{\mathrm{FM}}
+
(1-t)\lambda_\mathrm{ID}\mathcal{L}_\mathrm{ID},
\end{equation}
where $\lambda_\mathrm{ID}$ controls the contribution of the identity preservation term.

\subsubsection{Stage 2: restoration}

In Stage 2, we extend the training to full-parameter fine-tuning by unfreezing the DiT backbone $v_\theta$ together with the parameters optimized in Stage~1 and the input projection $W^{in'}$ presented in Eq.~\ref{eq:input_proj}. We illustrate this in Fig.~\ref{fig:pipeline}. At this stage, the identity conditioning is extended with conditioning on the actual degraded video $\tilde{I}^{1:K}$. The model is conditioned on the degraded video latent $\tilde{x}$, as described in Sec.~\ref{sec:deg_cond}, and is jointly optimized for identity preservation and perceptual fidelity. 

To encourage perceptual fidelity, we introduce a perceptual loss based on the DISTS metric for improved perceptual fidelity~\cite{ding2020image, osdface}. Following the same timestep-scaling rationale as in Stage 1, we introduce the Stage~2 objective as

\begin{equation} \label{eq:stage_2}
    \mathcal{L}_{\mathrm{Stage\text{-}2}} = \mathcal{L}_{\mathrm{FM}} + (1-t) \left( \lambda_{\mathrm{ID}} \mathcal{L}_{\mathrm{ID}} + \lambda_{\mathrm{DISTS}} \mathcal{L}_{\mathrm{DISTS}} \right) \text{,}
\end{equation} where $\lambda_{\mathrm{DISTS}}$ balances the contribution of the perceptual loss.

Taken together, our framework separates the sources of information required for identity-preserving restoration. The degraded video provides the spatial layout and motion of the target sequence, while the reference image provides bimodal identity representation that remains reliable even when the input frames are severely degraded. These cues are represented through complementary descriptive and perceptual features and injected into the pretrained flow-based text-to-video generator through parallel cross-attention. The two-stage training strategy first establishes effective identity conditioning before full restoration fine-tuning, reducing the tendency of the pretrained generative prior to favor perceptual realism over identity fidelity.

\section{Experimental setup}
\label{sec:experiments}

To validate the effectiveness of our approach, we conduct a series of experiments. We first describe the experimental setup, including datasets, implementation details, baselines, and ablation studies.

\subsection{Dataset details} \label{sec:dataset}

We train our model on VFHQ, a dataset of 15,000 high-quality face videos at 512$\times$512 resolution~\cite{vfhq}. Since VFHQ does not provide external reference images, we construct identity-matched references for each clip. Specifically, we select the frame with the highest MagFace~\cite{magface} face image quality score as the anchor frame and match it against candidate images from FFHQ~\cite{ffhq}, CelebA-HQ~\cite{celebahq}, CelebRef-HQ~\cite{dmdnet}, and additional web-collected references to obtain reference images $I^0$. Candidate images with identity loss, as defined in Eq.~\ref{eq:id_loss}, below 0.4 are retained as identity-matched references.

Following common practice in generative restoration, we synthesize clean-degraded video pairs during training using simulated blind degradation~\cite{wang2021realesrgan}. Given a high-quality video $I^{1:K}$, the degraded counterpart $\tilde{I}^{1:K}$ is generated as
\begin{equation}
    \tilde{I}^{1:K} = \left[(I^{1:K} * k_{\sigma_b}) \downarrow_s + n_{\sigma_n}\right]_{\mathrm{HEVC}_R},
    \label{eq:blind-degrad}
\end{equation}
where $k_{\sigma_b}$ denotes the blur kernel, $\downarrow_s$ denotes spatial downsampling by factor $s$, $n_{\sigma_n}$ denotes additive Gaussian noise, and $\mathrm{HEVC}_R$ denotes HEVC compression with constant rate factor $R$~\cite{sullivan2012overview}. During training, we sample $\sigma_b \sim \mathcal{U}(0.1,10)$ and $\sigma_n \sim \mathcal{U}(0,20)$, set the downsampling factor to $s=4$, and sample $R \sim \mathcal{U}(18,35)$.

We evaluate our model on the VFHQ-Test~\cite{vfhq}, which we extended, as described above, to include reference frames. We report results under validation degradation setting $\sigma_b \in [5.0,10.0]$, $\sigma_n \in [5.0,10.0]$, fixed downsampling factor $s=4$, and $Q \in [25,35]$. Pixel-level fidelity is measured via PSNR and SSIM, perceptual fidelity via LPIPS~\cite{lpips} and DISTS~\cite{dists}, identity fidelity via the ArcFace identity similarity score (IDS)~\cite{arcface}, temporal identity stability via VIDD~\cite{vidd}, and temporal realism via FVD~\cite{fvd}. We report further results under different degradation parameters in the appendix.

\subsection{Baselines}

We compare our method against four categories of restoration approaches. The first group includes reference-free face image restoration methods: CodeFormer~\cite{codeformer}, DifFace~\cite{difface}, and DiffBIR~\cite{diffbir}. The second group contains reference-guided face image restoration methods: ReF-LDM~\cite{refldm}, RestorerID~\cite{restorerid}, and InstantRestore~\cite{instantrestore}. The third group consists of reference-free face video restoration methods: KEEP~\cite{keep0}, PGTFormer~\cite{pgtformer}, and SVFR~\cite{svfr}. Finally, we include SeedVR-3B~\cite{seedvr0}, a generic video restoration model based on a large-scale diffusion backbone. To the best of our knowledge, there is no directly comparable subject-agnostic reference-guided method for face video restoration. We use the publicly available models and source code released by the authors, applying image-based methods frame-by-frame and video-based methods to the full input sequence. All methods are evaluated on the same degraded inputs and compared against the corresponding clean videos from the test set described above.

\subsection{Implementation Details}

We use Wan2.1-T2V-1.3B as the pretrained generative backbone, which consists of a 3D causal variational autoencoder (VAE) used as $\mathcal{E}$ and $\mathcal{D}$, a T5 transformer used as the text encoder $\mathcal{E}_\mathrm{T}$, and a DiT backbone used as the velocity model $v_\theta$~\cite{wan0}. We use ArcFace as the face encoder $\mathcal{E}_\mathrm{ID}$~\cite{arcface}. As described in Sec.~\ref{sec:stage_1}, Stage~1 trains only the introduced identity-conditioning parameters. We use the AdamW optimizer with a learning rate of $8{\times}10^{-5}$, a cosine decay schedule with 250 warmup steps, a per-device batch size of $2$, and $4$ gradient accumulation steps. We set $\lambda_{\mathrm{ID}}=1.0$ and train for $13{,}500$ steps on $3\times$ NVIDIA RTX Pro 6000 Blackwell GPUs.

In Stage~2, we jointly fine-tune the introduced BIR, PDC, and the backbone parameters $v_\theta$, for $30{,}000$ steps with $\lambda_{\mathrm{ID}}=1.0$ and $\lambda_{\mathrm{DISTS}}=1.0$. The learning rate is reduced to $2{\times}10^{-5}$ and the number of gradient accumulation steps is increased to $8$, while the cosine schedule, warmup configuration, batch size, and hardware setup remain unchanged. In both stages, we drop perceptual and descriptive identity features with probability $0.1$ for classifier-free guidance~\cite{cfg}. During inference, we use $50$ sampling steps and a classifier-free guidance scale of $w=2.0$.

The identity and perceptual losses introduced in Eqs.~\ref{eq:id_loss} and~\ref{eq:stage_2} require estimating the clean latent $\hat{x}_{t\rightarrow0}$ and decoding it to image space as $\mathcal{D}(\hat{x}_{t\rightarrow0})$. To reduce computational cost, we compute these losses only on the first frame of each video during training. This is compatible with the causal VAE design of Wan2.1, in which the first frame provides a representation of all subsequent frames.

\subsection{Ablation studies}

We conduct ablation studies to analyze the contribution of the main components of our method. Following the two-stage training procedure described in Sec.~\ref{sec:training_stages}, we evaluate the identity-conditioning stage and the restoration stage separately. This allows us to isolate the effects of descriptive identity and perceptual identity conditioning, as well as perceptual restoration supervision.

For Stage~1, we first study the effect of the identity loss weight $\lambda_{\mathrm{ID}}$ in Eq.~\ref{eq:stage_1_loss}. We then evaluate the contribution of the local and global perceptual features, $c_p^{local}$ and $c_p^{global}$, introduced in the BIR module in Sec.~\ref{sec:bir}. Finally, we assess the effectiveness of PDC by ablating the perceptual and descriptive conditioning signals, $c_p$ and $c_d$, described in Sec.~\ref{sec:pdc}.

For Stage~2, we evaluate the effect of the perceptual loss $\mathcal{L}_{\mathrm{DISTS}}$. We further test the importance of the two-stage training strategy by comparing it against single-stage full fine-tuning. Finally, we disable our introduced modules BIR and PDC, \ie, identity conditioning, and finetune only the backbone DiT model $v_\theta$ in a single stage.

All ablations are conducted on a separate evaluation split constructed from HDTF~\cite{hdtf}, consisting of 50 subjects. This split is disjoint from the test set used for baseline comparisons, preventing ablation-driven parameter selection on the main evaluation benchmark.

\input{tables/quantitative_results_medium}

\section{Results \& Discussion}

\subsection{Quantitative results}

Table~\ref{tab:quantitative_results_medium_ref_based} reports quantitative results on the test set using the metrics described in Sec.~\ref{sec:dataset}. Our method achieves the best performance in PSNR, LPIPS, DISTS, IDS, and FVD, demonstrating strong reconstruction fidelity, perceptual quality, identity preservation, and video realism. The improvement in IDS is particularly notable, even compared with reference-guided image restoration baselines, highlighting the effectiveness of the proposed identity conditioning. InstantRestore obtains the second-best IDS score, but as an image-based method, it does not explicitly enforce temporal consistency. In contrast, our method achieves the best FVD and competitive VIDD, indicating improved video quality while maintaining stable identity across frames. Although some baselines achieve slightly higher SSIM or VIDD, they do so at the expense of substantially weaker identity preservation or perceptual quality.

\subsection{Qualitative results}

\begin{figure*}[htbp!]
\begin{center}
\includegraphics[width=\textwidth]{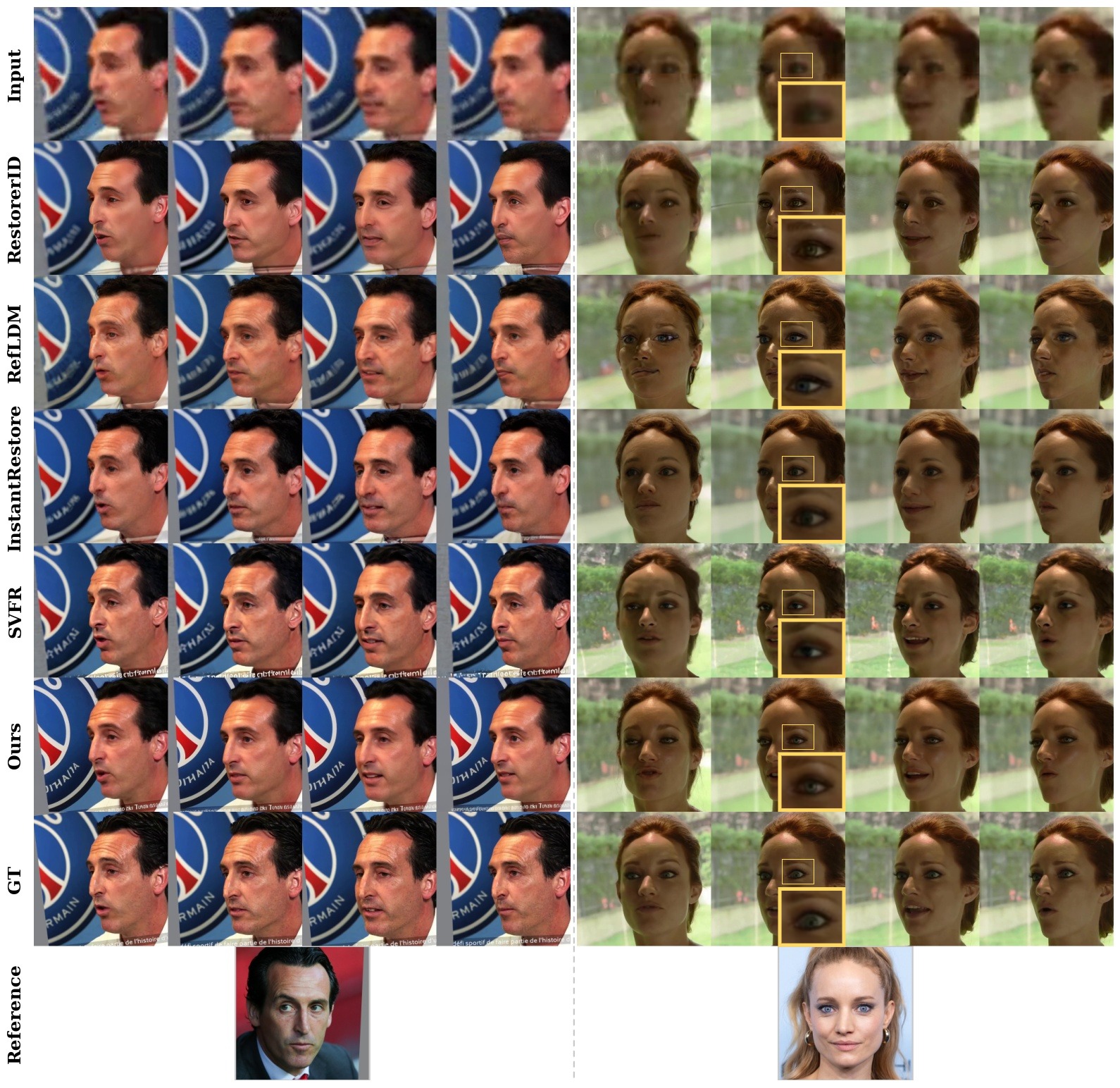}
\end{center}
   \caption{Qualitative results. The rows display the degraded input, the baselines, our method, and the ground-truth, with the reference images at the bottom.}
\label{fig:grid_comparison}
\end{figure*}

\begin{figure*}[htbp!]
\begin{center}
\begin{tabular}{c@{\hspace{0.001\textwidth}}c}
\bmvaHangBox{\includegraphics[width=0.495\textwidth]{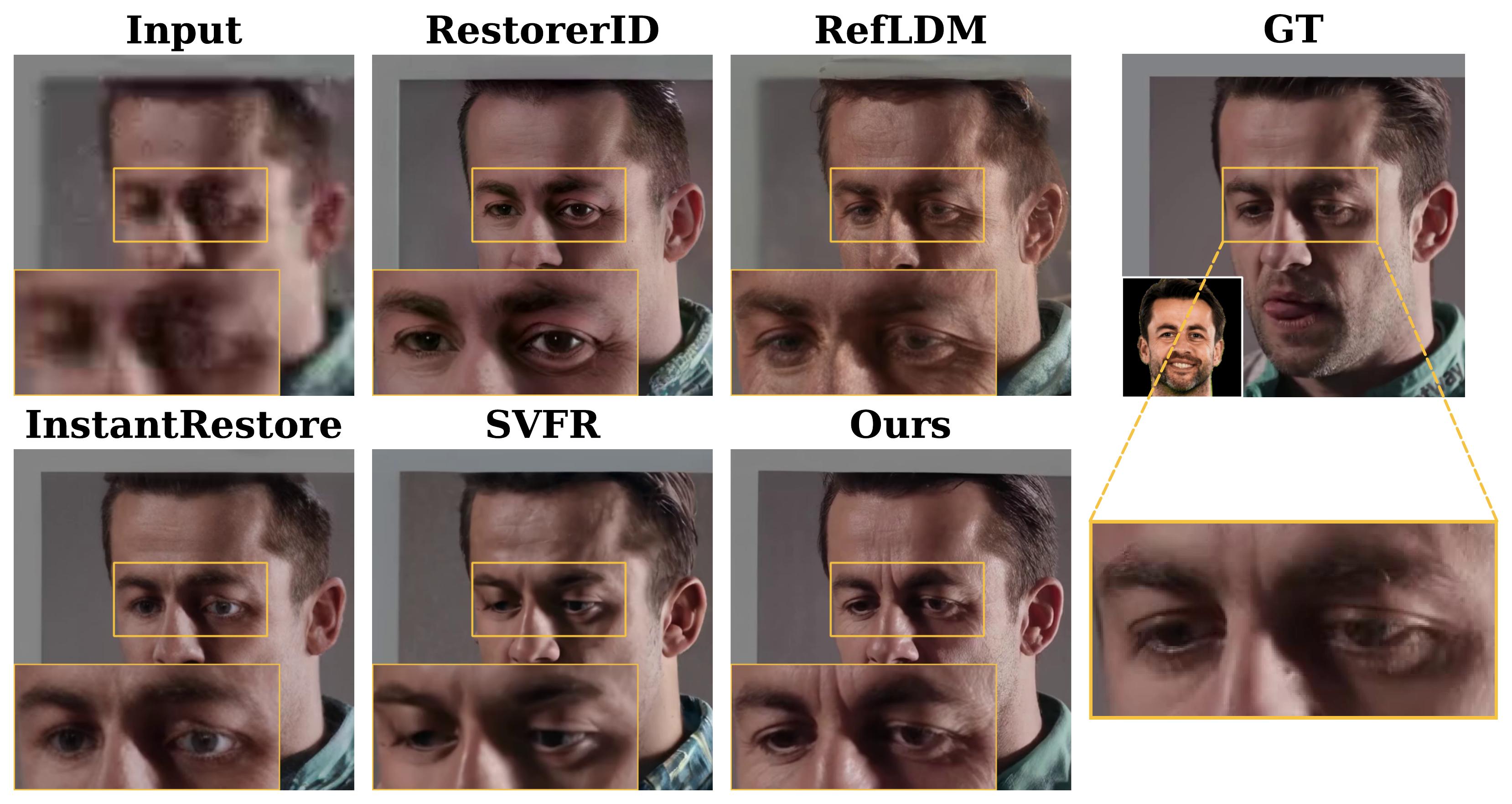}} &
\bmvaHangBox{\includegraphics[width=0.495\textwidth]{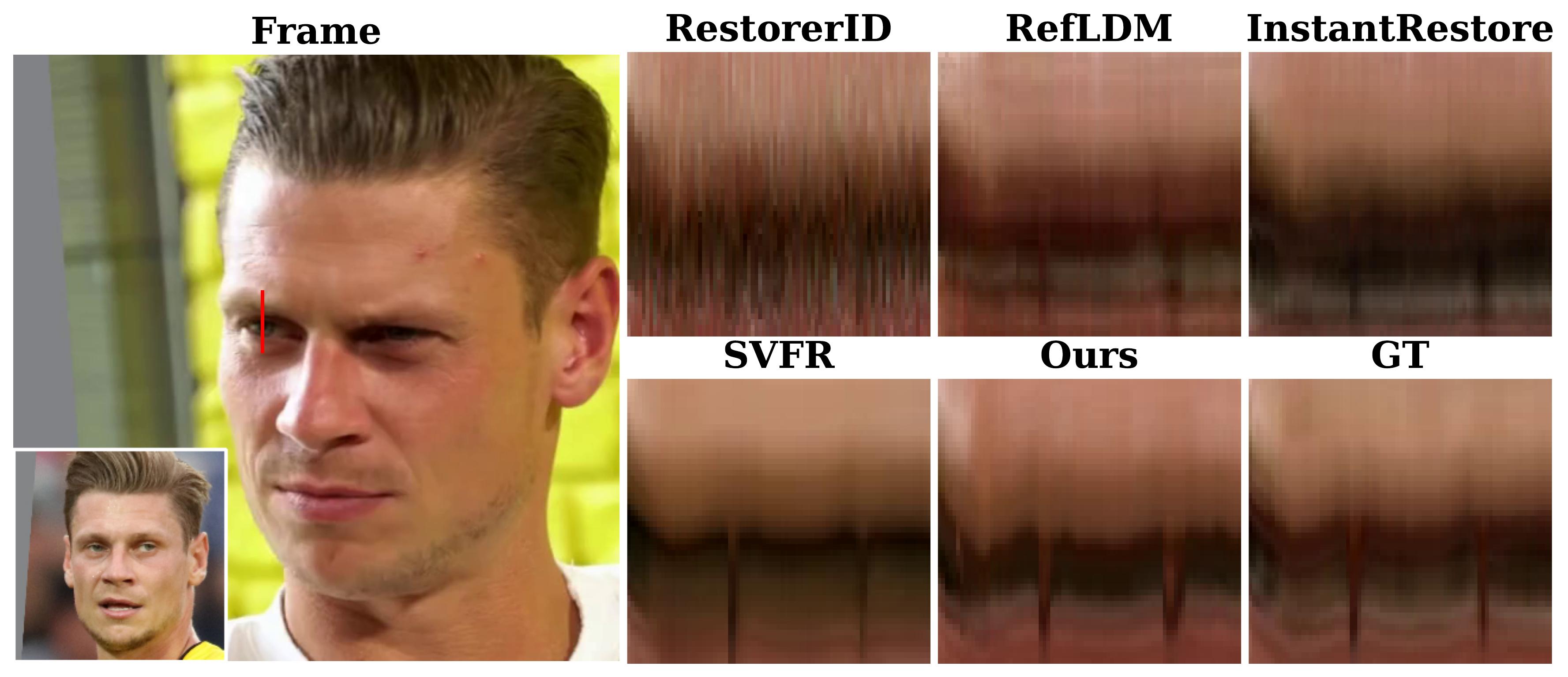}} \\
(a) Comparing identity preservation. &
(b) Comparing temporal consistency.
\end{tabular}
\end{center}
    \caption{Qualitative comparisons of identity preservation and temporal consistency.}
\label{fig:qual2}
\end{figure*}

Figure~\ref{fig:grid_comparison} presents a qualitative comparison of examples from the test set. Here, we include reference-guided baselines together with the reference-free video restoration method SVFR due to space constraints. We provide additional videos and methods in the appendix. We observe that image-based reference-guided methods are sensitive to pose and expression differences between the degraded frame and the reference image, often leading to distorted facial structure and suboptimal identity preservation. Furthermore, these methods suffer from temporal inconsistencies. SVFR, by contrast, benefits from temporal modeling and therefore produces temporally stable, visually plausible results. However, a lack of identity preservation leads to inconsistencies; for example, it cannot preserve the blue eyes of the right-hand subject in Fig.~\ref{fig:grid_comparison}. Our method, on the other hand, enables restoration with temporal consistency while preserving the subject's identity without requiring any subject-specific fine-tuning. We further present in Fig.~\ref{fig:qual2} (a) a close-up qualitative comparison and in Fig.~\ref{fig:qual2} (b) a temporal consistency comparison.

\subsection{Ablation studies}

We divide the ablation study into two stages. The first stage focuses on identity preservation and therefore reports IDS and VIDD. As shown in Tab.~\ref{tab:stage1_ablations}, increasing $\lambda_{\mathrm{ID}}$ improves both identity similarity (IDS) and temporal identity consistency (VIDD) within the tested range. We therefore set $\lambda_{\mathrm{ID}}=1.0$ for all experiments. We then ablate the feature global-local composition of $c_p$. The results show that combining global and local perceptual features, $c_p^{global} + c_p^{local}$, provides the best identity-preservation performance among the tested variants. Removing $c_d$ degrades both IDS and VIDD, suggesting that the bimodal features extracted through BIR (Sec.~\ref{sec:bir}), together with $\mathcal{L}_\mathrm{ID}$, contribute to identity preservation.

The second-stage ablations focus on both identity preservation and restoration quality. Tab.~\ref{tab:stage2_ablations} shows that adding $\mathcal{L}_\mathrm{DISTS}$ improves perceptual quality while also benefiting identity preservation. The two-stage training strategy further improves identity-related metrics, supporting our hypothesis that single-stage training can weaken the effect of identity-conditioning features. We finally show in Tab.~\ref{tab:full_ablation} that removing the proposed BIR and PDC modules and training only $v_\theta$ for restoration leads to substantially worse performance. This suggests that the gains do not trivially come from the backbone.

\input{tables/ablations}

\section{Conclusion}

We introduced a subject-agnostic, reference-guided method for face video restoration that improves pixel-level, perceptual, and identity fidelity. Our method extends a large-scale pretrained flow-matching backbone with bimodal identity representations from BIR and conditioning through PDC. By jointly modeling bimodal identity features, combining local-global perceptual and attribute-based descriptive features, and adopting a two-stage training strategy, our approach improves identity preservation and restoration quality under challenging identity-obscuring degradations. The ablation studies further demonstrated the contribution of the proposed components.

Our method has several limitations. First, it is inefficient due to the iterative nature of flow matching. Second, its advantage over existing baselines becomes less pronounced when the input video is not heavily degraded. Finally, our evaluation is limited to a specific set of degradations. Future work will explore broader identity-obscuring degradations, including fog, low light, haze, and occlusion, and investigate more efficient architectures and flow-matching step-reduction techniques.

\bibliography{main}

\newpage
\appendix
\input{appendix.tex}

\end{document}

%% file: tables/quantitative_results_medium.tex
\begin{table}[t!]
\begin{center}
\small{
\begin{tabular}{lccccccc}
\toprule
% \multicolumn{8}{c}{\textbf{VFHQ-Ref-Test -- Medium Degradation}} \\
% \midrule
\multicolumn{1}{c}{Methods} & PSNR$\uparrow$ & SSIM$\uparrow$ & LPIPS$\downarrow$ & DISTS$\downarrow$ & IDS$\uparrow$ & VIDD$\downarrow$ & FVD$\downarrow$ \\
\midrule
CodeFormer~\cite{codeformer} & 25.293 & 0.769 & 0.262 & 0.160 & 0.515 & 0.429 & 430.62 \\
DifFace~\cite{difface} & 25.233 & 0.769 & 0.285 & 0.166 & 0.374 & 1.062 & 2301.76 \\
DiffBIR~\cite{diffbir} & 25.277 & 0.733 & 0.325 & 0.186 & 0.563 & 0.708 & 878.83 \\
%OSDFace~\cite{osdface} & 23.483 & 0.712 & 0.269 & 0.154 & 0.592 & 0.392 & 432.47 \\
RestorerID~\cite{restorerid} & 24.970 & 0.748 & 0.300 & 0.171 & 0.581 & 0.752 & 1193.27 \\
Ref-LDM~\cite{refldm} & 24.841 & 0.748 & 0.290 & 0.165 & \third{0.612} & 0.383 & 410.93 \\
InstantRestore~\cite{instantrestore} & 25.884 & 0.794 & \third{0.217} & 0.150 & \second{0.707} & 0.318 & 346.98 \\
\midrule
SeedVR-3B~\cite{seedvr0} & 24.488 & 0.776 & 0.250 & \third{0.149} & 0.592 & \third{0.256} & \third{290.21} \\
PGTFormer~\cite{pgtformer} & \third{26.104} & 0.794 & 0.266 & 0.167 & 0.557 & 0.351 & 480.31 \\
KEEP~\cite{keep0} & \second{26.211} & \best{0.807} & \second{0.210} & 0.151 & 0.545 & 0.347 & 409.60 \\
SVFR~\cite{svfr} & 25.533 & \second{0.796} & 0.225 & \second{0.138} & 0.548 & \best{0.191} & \second{229.21} \\
Ours & \best{26.231} & \third{0.795} & \best{0.202} & \best{0.134} & \best{0.711} & \second{0.209} & \best{208.73} \\
%\midrule
%Ours ($v_\theta \text{ only}$) & 24.277 & 0.755 & 0.300 & 0.179 & 0.386 & 0.221 & 479.52 \\
\bottomrule
\end{tabular}%
}
\end{center}
\caption{Quantitative comparison with state-of-the-art methods. \best{Best}, \second{second-best}, and \third{third-best} results are highlighted. Our method achieves the best pixel, perceptual, and identity fidelity performance.}
\label{tab:quantitative_results_medium_ref_based}
\end{table}

%% file: tables/ablations.tex
\begin{table}[t!]
\small
\setlength{\tabcolsep}{4pt}
\begin{center}
\begin{tabular}{@{}ccc|ccccc@{}}
\toprule
$\lambda_{\mathrm{ID}}$ & IDS $\uparrow$ & VIDD $\downarrow$
& $c_p^{local}$ & $c_p^{global}$ & $c_d$ & IDS $\uparrow$ & VIDD $\downarrow$ \\
\midrule
0.0  & 0.151         & 0.442         & \textbf{\checkmark} & \ding{55}            & \textbf{\checkmark} & \third{0.369}  & 0.369 \\
0.25 & \third{0.394} & \third{0.395} & \ding{55}            & \textbf{\checkmark} & \textbf{\checkmark} & 0.367          & \best{0.242} \\
0.5  & \second{0.447}& \second{0.331}& \textbf{\checkmark} & \textbf{\checkmark} & \ding{55}            & \second{0.406} & \third{0.338} \\
1.0  & \best{0.481}  & \best{0.282}  & \textbf{\checkmark} & \textbf{\checkmark} & \textbf{\checkmark} & \best{0.481}   & \second{0.282} \\
\bottomrule
\end{tabular}
\end{center}
\caption{Stage 1 ablation studies. $\lambda_{ID}=1.0$ and enabling both local-global perceptual conditioning and descriptive conditioning yield the best identity preservation.}
\label{tab:stage1_ablations}
\end{table}

\begin{table}[t!]
\begin{center}
\small
\setlength{\tabcolsep}{4pt}
\begin{tabular}{@{}cc|ccccccc@{}}
\toprule
$\mathcal{L}_{\mathrm{DISTS}}$ & Two Stage
& PSNR $\uparrow$ & SSIM $\uparrow$ & LPIPS $\downarrow$ & DISTS $\downarrow$ & IDS $\uparrow$ & VIDD $\downarrow$ & FVD $\downarrow$ \\
\midrule
\ding{55}            & \textemdash      & \best{23.305}  & \best{0.7480} & \second{0.271} & \second{0.157} & \second{0.547} & \best{0.210}   & \best{329.18} \\
\textbf{\checkmark} & \textemdash      & \second{23.251}& \second{0.742} & \best{0.271}   & \best{0.156}   & \best{0.551}   & \second{0.214} & \second{345.46} \\
\midrule
\textemdash          & \ding{55}        & \best{23.965}  & \best{0.772}  & \best{0.266}   & \second{0.163} & \second{0.435} & \second{0.234} & \second{364.09} \\
\textemdash          & \textbf{\checkmark} & \second{23.251}& \second{0.742} & \second{0.271} & \best{0.156}   & \best{0.551}   & \best{0.214}   & \best{345.46} \\
\bottomrule
\end{tabular}
\end{center}
\caption{Stage 2 ablation studies. $\mathcal{L}_\mathrm{DISTS}$ enables perceptual fidelity and two-stage training improves identity preservation.}
\label{tab:stage2_ablations}
\end{table}

\begin{table}[t!]
\begin{center}
\small
\setlength{\tabcolsep}{4pt}
\begin{tabular}{@{}c|ccccccc@{}}
\toprule
Model & PSNR $\uparrow$ & SSIM $\downarrow$ & LPIPS $\downarrow$ & DISTS $\downarrow$ & IDS $\uparrow$ & VIDD $\downarrow$ & FVD $\downarrow$ \\ \midrule
$v_\theta$ only & \best{24.277} & \best{0.755} & 0.300 & 0.179 & 0.386 & 0.221 & 479.52 \\
Full & 23.251 & 0.742 & \best{0.271} & \best{0.156} & \best{0.551} & \best{0.214} & \best{345.46} \\ \bottomrule
\end{tabular}
\end{center}
\caption{Ablating backbone-only training. Our method overall improves perceptual and identity fidelity as well as the temporal consistency.}
\label{tab:full_ablation}
\end{table}

%% file: appendix.tex
\setcounter{page}{1}

\section{Additional Qualitative Results}

\begin{figure*}[htbp!]
    \begin{center}
    \includegraphics[width=0.85\textwidth]{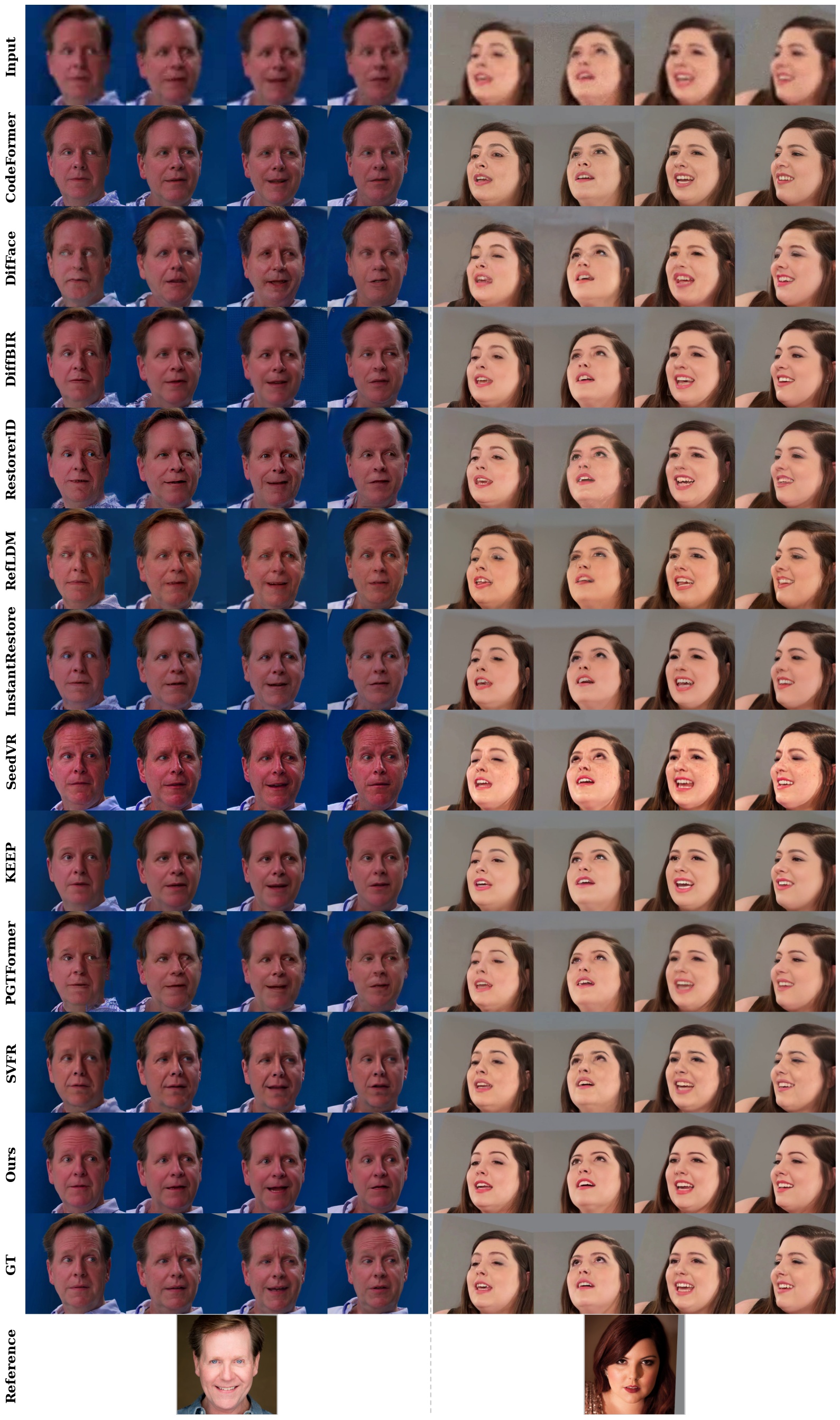}
    \end{center}
    \label{fig:grid_comparison_medium1}
\end{figure*}

\begin{figure*}[htbp!]
    \begin{center}
    \includegraphics[width=0.85\textwidth]{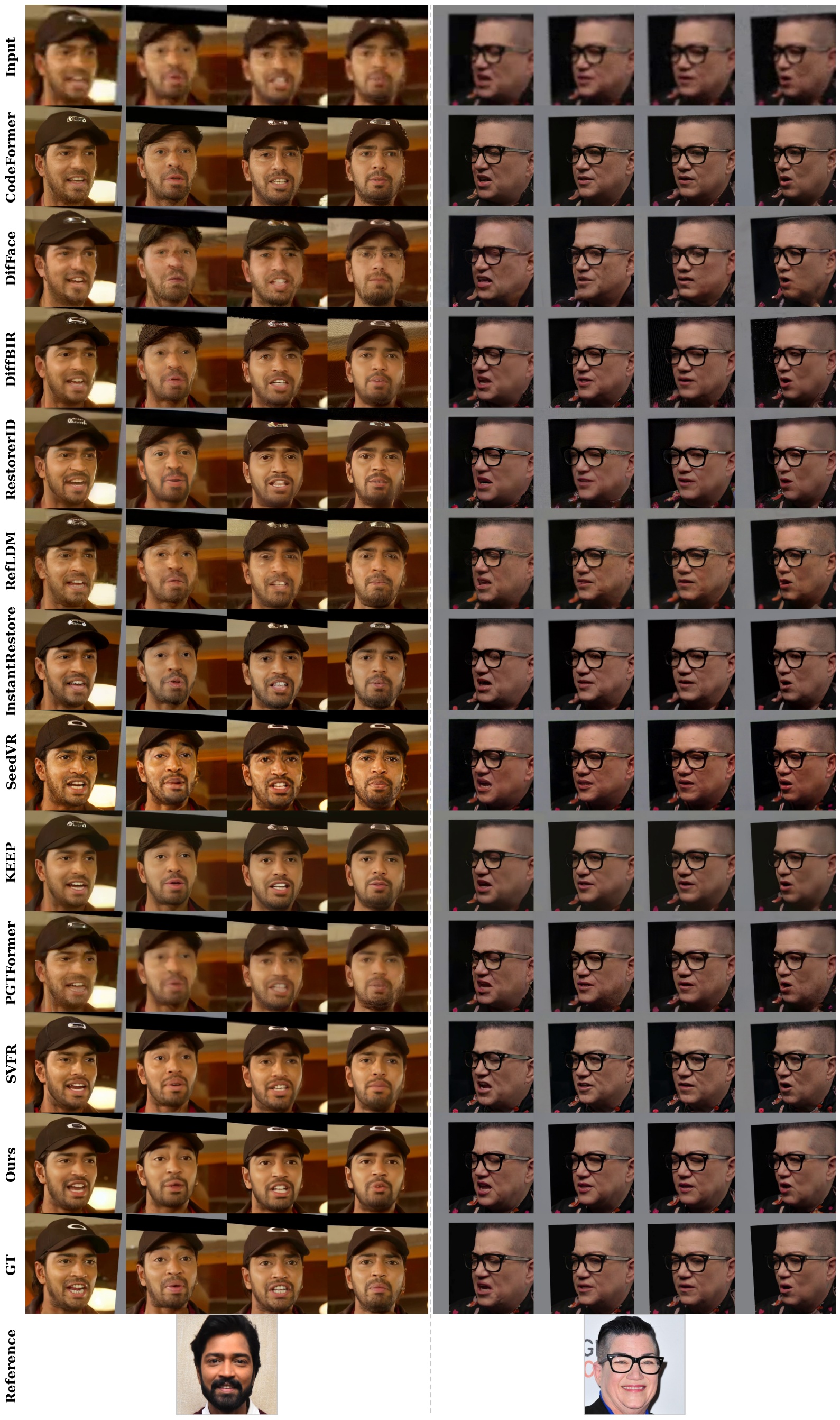}
    \end{center}
    \label{fig:grid_comparison_medium2}
\end{figure*}

\begin{figure*}[htbp!]
    \begin{center}
    \includegraphics[width=0.85\textwidth]{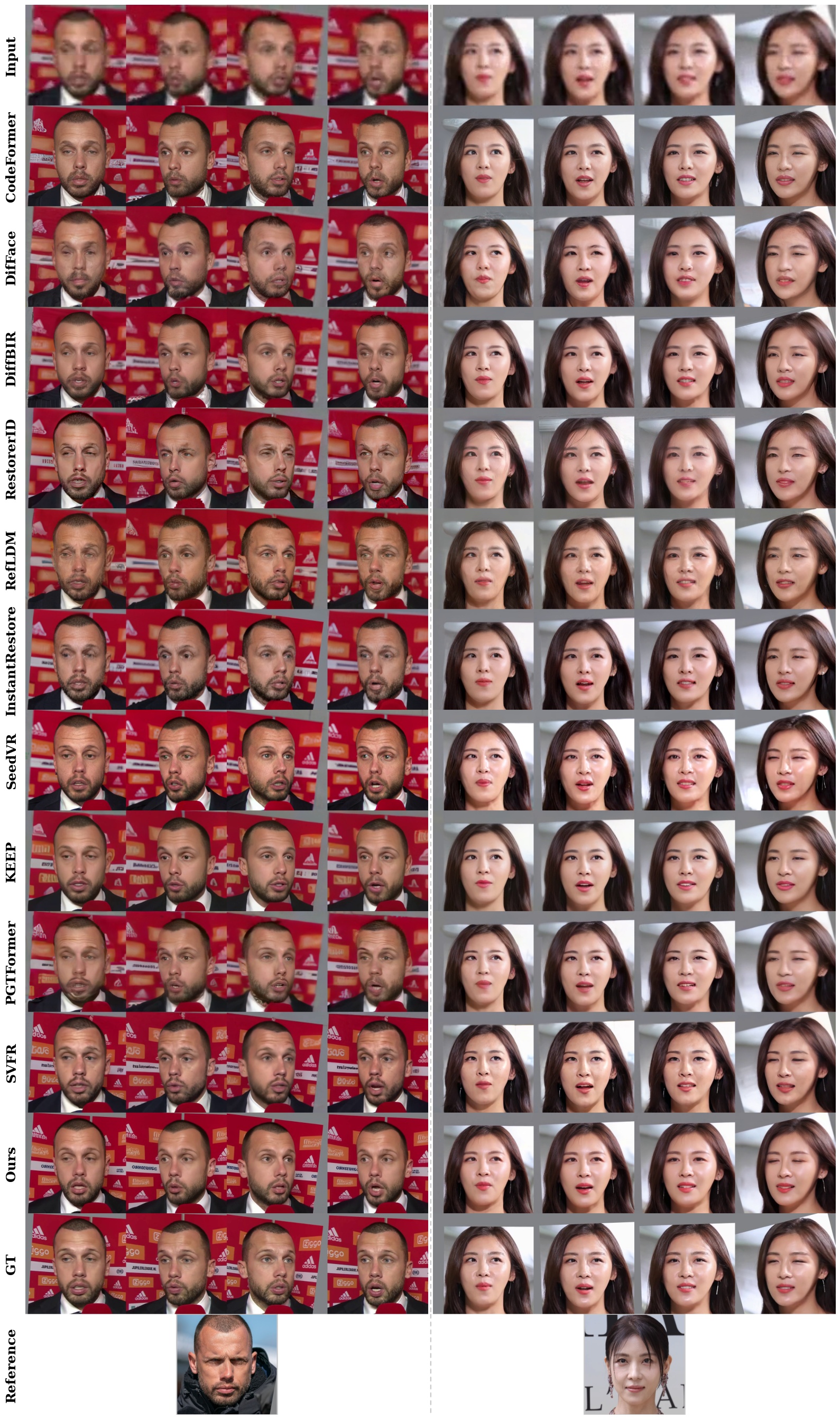}
    \end{center}
    \label{fig:grid_comparison_medium3}
\end{figure*}

\newpage
\section{Additional Degradation Levels: Milder}

\vspace*{\fill}
\input{tables/quantitative_results_mild}
\vspace*{\fill}

\newpage

\begin{figure*}[htbp!]
    \begin{center}
    \includegraphics[width=0.85\textwidth]{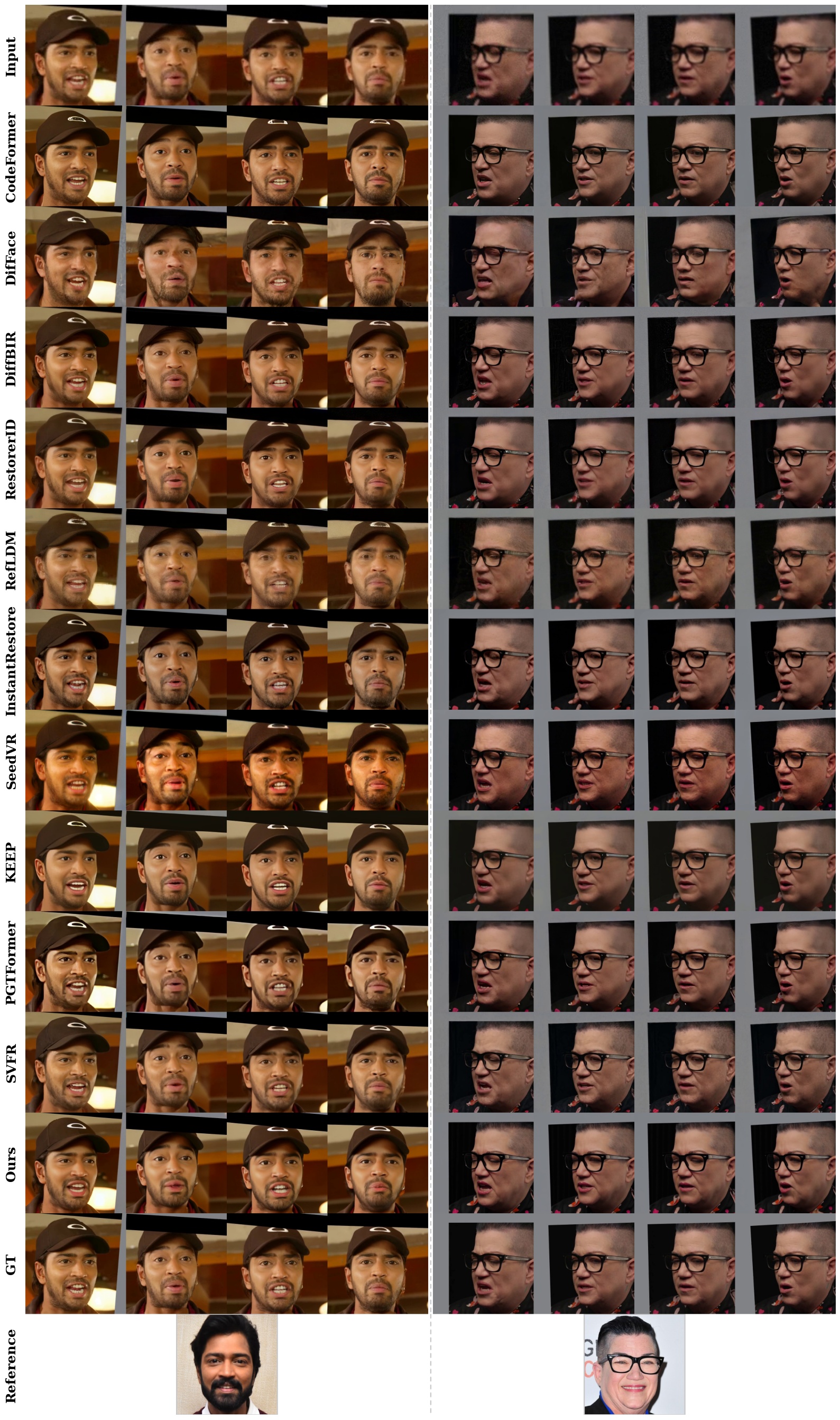}
    \end{center}
    \label{fig:grid_comparison_mild1}
\end{figure*}

\begin{figure*}[htbp!]
    \begin{center}
    \includegraphics[width=0.85\textwidth]{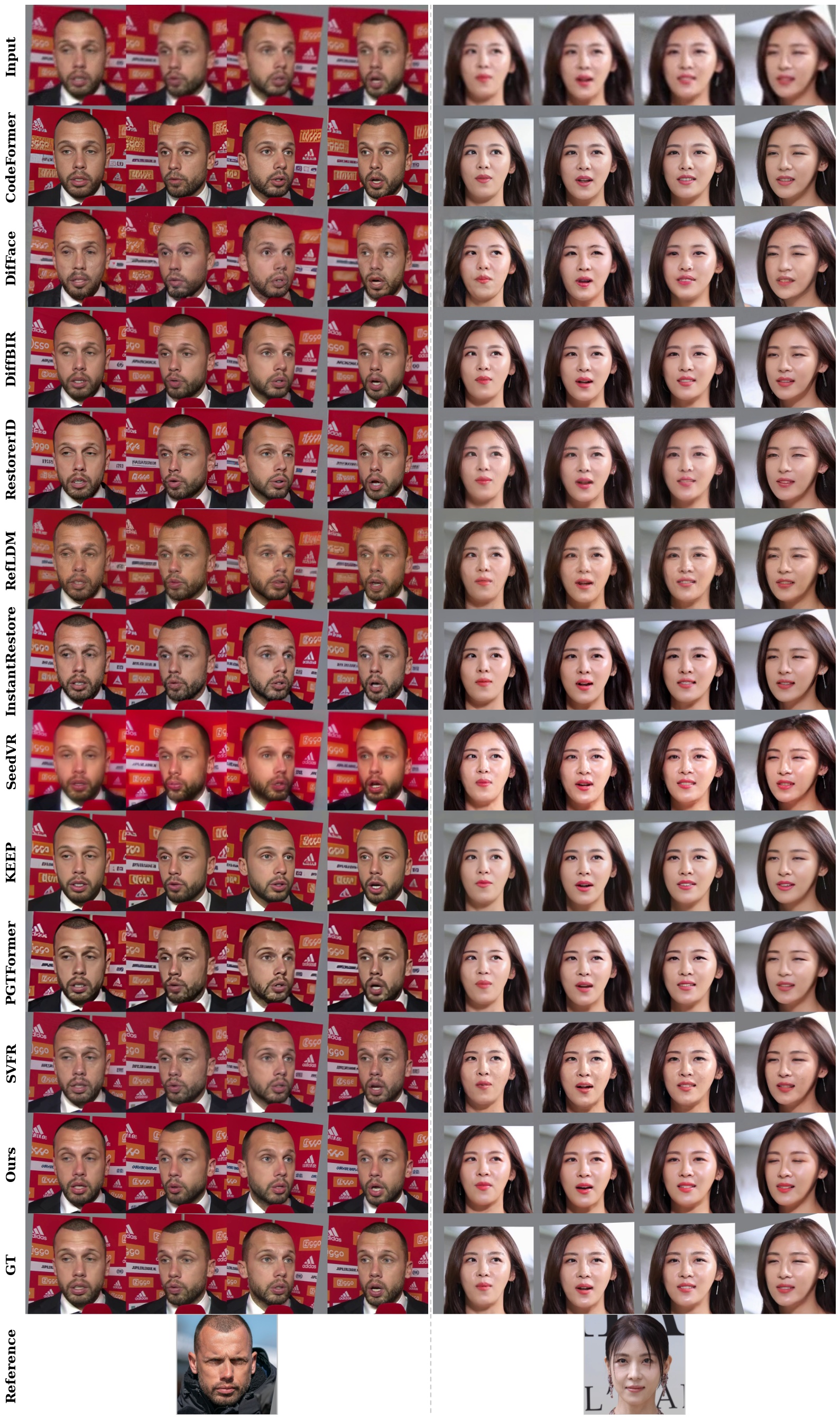}
    \end{center}
    \label{fig:grid_comparison_mild2}
\end{figure*}

\section{Additional Degradation Levels: Heavier}

\vspace*{\fill}
\input{tables/quantitative_results_severe}
\vspace*{\fill}

\begin{figure*}[htbp!]
    \begin{center}
    \includegraphics[width=0.85\textwidth]{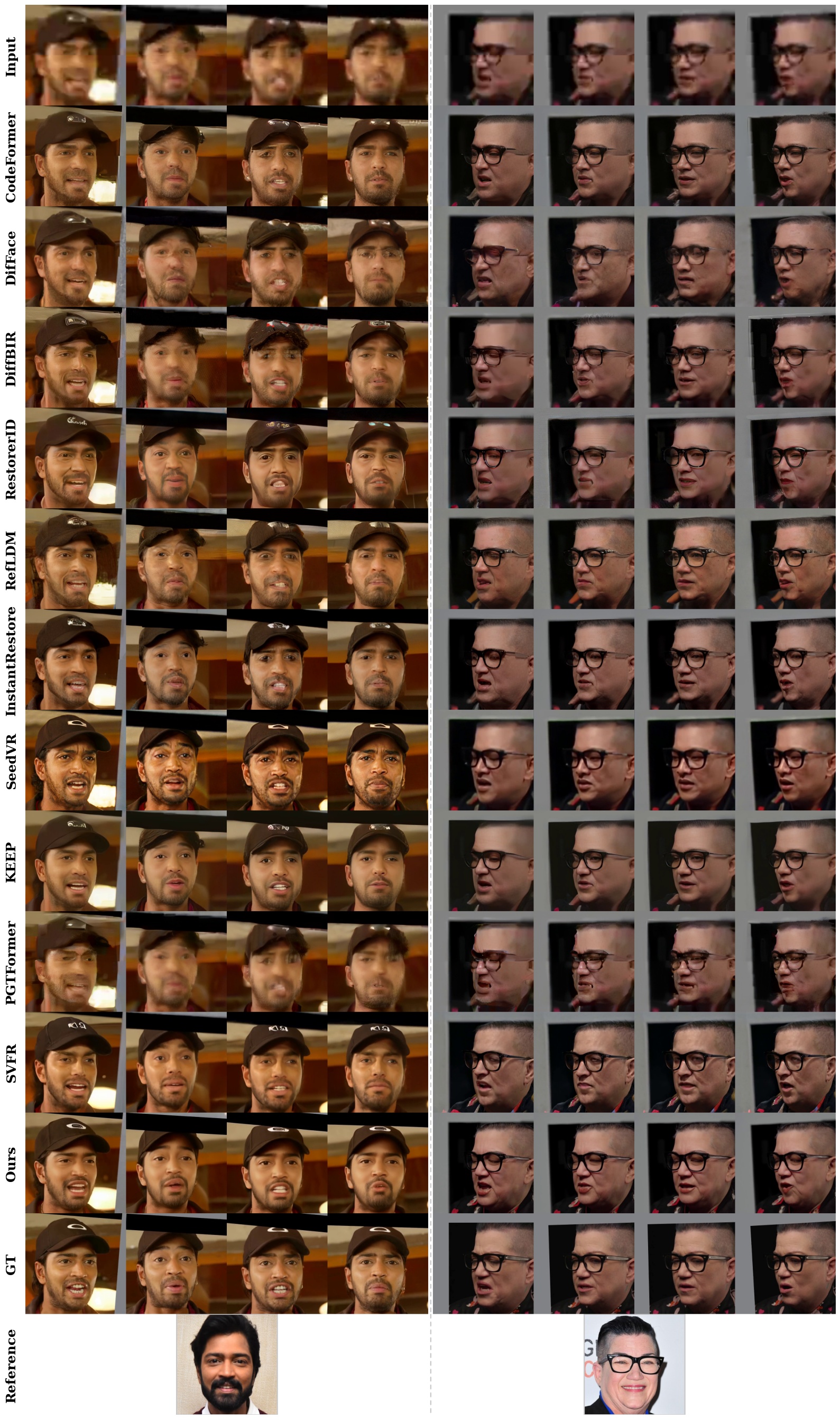}
    \end{center}
    \label{fig:grid_comparison_hard1}
\end{figure*}

\begin{figure*}[htbp!]
    \begin{center}
    \includegraphics[width=0.85\textwidth]{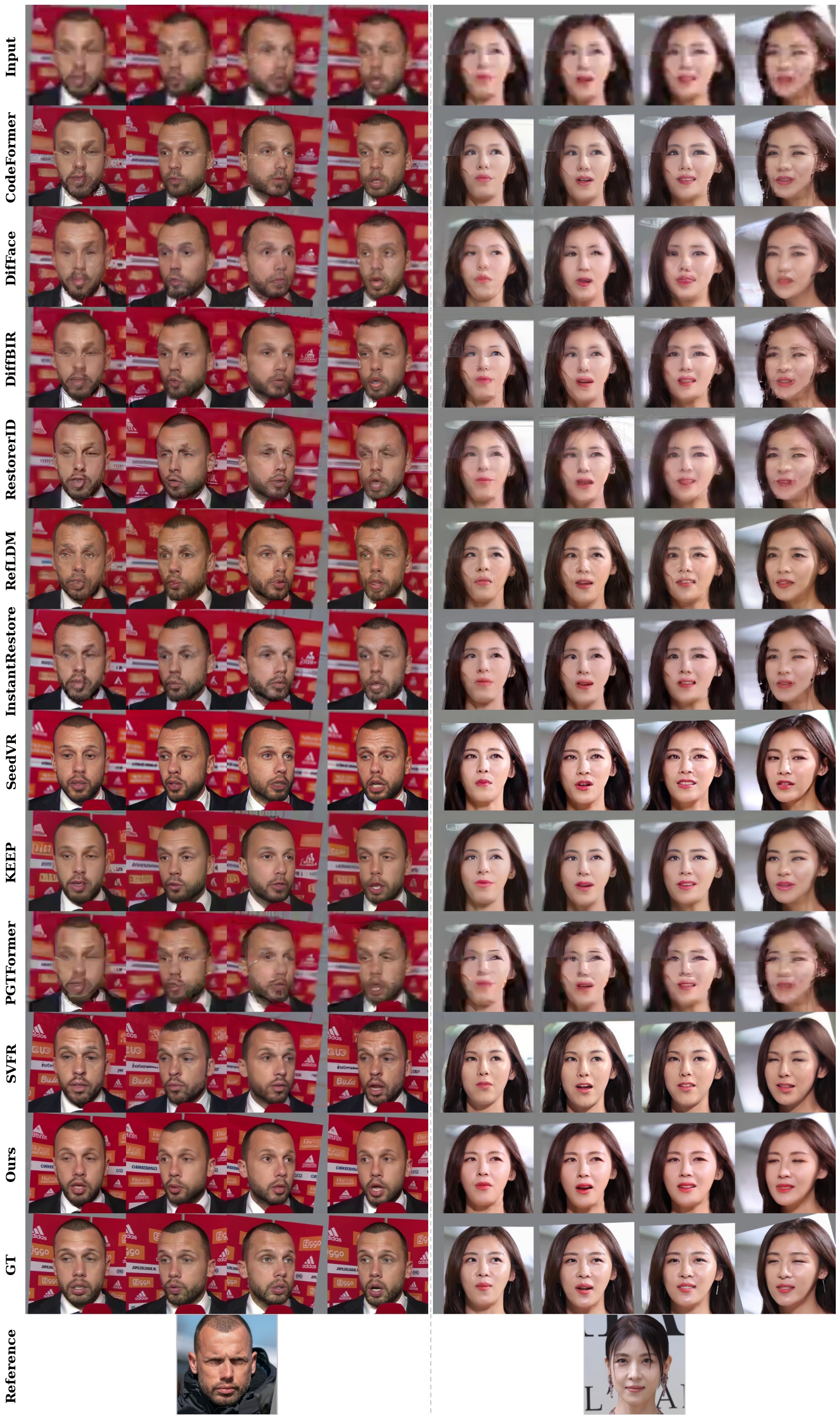}
    \end{center}
    \label{fig:grid_comparison_hard2}
\end{figure*}

%% file: tables/quantitative_results_mild.tex
\begin{table*}[htbp!]
\begin{center}
\footnotesize{
\begin{tabular}{lccccccc}
\toprule
\multicolumn{8}{c}{$\sigma_b \in [2,5]$, $s=4$, $\sigma_n \in [5, 10]$, and $R\in [18,25]$} \\
\midrule
\multicolumn{1}{c}{Methods} & PSNR$\uparrow$ & SSIM$\uparrow$ & LPIPS$\downarrow$ & DISTS$\downarrow$ & IDS$\uparrow$ & VIDD$\downarrow$ & FVD$\downarrow$ \\
\midrule
CodeFormer~\cite{codeformer} & 27.308 & 0.820 & 0.184 & 0.130 & 0.743 & 0.357 & 203.24 \\
DifFace~\cite{difface} & 26.475 & 0.789 & 0.241 & 0.143 & 0.498 & 1.019 & 1975.87 \\
DiffBIR~\cite{diffbir} & 27.685 & 0.785 & 0.234 & 0.153 & 0.832 & 0.522 & 299.37 \\
%OSDFace~\cite{osdface} & 24.842 & 0.751 & 0.232 & 0.145 & 0.801 & 0.347 & 271.72 \\
RestorerID~\cite{restorerid} & 27.591 & 0.806 & 0.208 & 0.139 & 0.813 & 0.553 & 403.13 \\
ReF-LDM~\cite{refldm} & 26.710 & 0.796 & 0.234 & 0.146 & 0.790 & 0.377 & 231.40 \\
InstantRestore~\cite{instantrestore} & 27.734 & 0.831 & 0.164 & 0.129 & \third{0.842} & 0.288 & 194.61 \\
\midrule
SeedVR-3B~\cite{seedvr0} & 26.138 & 0.825 & 0.182 & 0.125 & 0.839 & 0.269 & 212.82 \\
PGTFormer~\cite{pgtformer} & \best{28.945} & \best{0.855} & \best{0.150} & \second{0.117} & \best{0.861} & \third{0.256} & \third{160.93} \\
KEEP~\cite{keep0} & \third{28.106} & \third{0.845} & \third{0.158} & 0.126 & 0.764 & 0.299 & 246.20 \\
SVFR~\cite{svfr} & \second{28.736} & \second{0.852} & \second{0.151} & \best{0.112} & \second{0.855} & \best{0.185} & \best{110.98} \\
Ours & 27.657 & 0.825 & 0.161 & \third{0.118} & 0.834 & \second{0.201} & \second{131.33} \\
\bottomrule
\end{tabular}%
}
\end{center}
\caption{Quantitative comparison with state-of-the-art methods under milder test degradation settings. \best{Best}, \second{second-best}, and \third{third-best} results are highlighted.}
\end{table*}

%% file: tables/quantitative_results_severe.tex
\begin{table*}[htbp!]
\begin{center}
\footnotesize{
\begin{tabular}{lccccccc}
\toprule
\multicolumn{8}{c}{$\sigma_b \in [5, 10]$, $s=4$, $\sigma_n \in [5, 10]$, and $R\in [35, 45]$} \\
\midrule
\multicolumn{1}{c}{Methods} & PSNR$\uparrow$ & SSIM$\uparrow$ & LPIPS$\downarrow$ & DISTS$\downarrow$ & IDS$\uparrow$ & VIDD$\downarrow$ & FVD$\downarrow$ \\
\midrule
CodeFormer~\cite{codeformer} & 23.479 & 0.736 & 0.335 & 0.186 & 0.325 & 0.434 & 748.81 \\
DifFace~\cite{difface} & 23.578 & 0.743 & 0.340 & 0.193 & 0.255 & 1.082 & 2741.15 \\
DiffBIR~\cite{diffbir} & 23.290 & 0.689 & 0.437 & 0.241 & 0.323 & 0.774 & 1389.80 \\
%OSDFace~\cite{osdface} & 21.920 & 0.681 & 0.324 & \third{0.169} & 0.388 & 0.380 & 616.77 \\
RestorerID~\cite{restorerid} & 23.238 & 0.718 & 0.369 & 0.197 & 0.407 & 0.837 & 1729.56 \\
ReF-LDM~\cite{refldm} & 23.113 & 0.714 & 0.341 & 0.179 & \third{0.496} & 0.346 & \third{541.39} \\
InstantRestore~\cite{instantrestore} & \third{23.909} & \third{0.762} & \third{0.280} & 0.170 & \second{0.553} & 0.323 & 627.65 \\
\midrule
SeedVR-3B~\cite{seedvr0} & 22.818 & 0.753 & 0.326 & 0.189 & 0.356 & \third{0.250} & 543.17 \\
PGTFormer~\cite{pgtformer} & 23.591 & 0.753 & 0.377 & 0.212 & 0.262 & 0.342 & 1239.43 \\
KEEP~\cite{keep0} & \second{24.125} & \best{0.774} & \second{0.270} & 0.171 & 0.362 & 0.326 & 687.610 \\
SVFR~\cite{svfr} & 23.166 & 0.759 & 0.288 & \second{0.160} & 0.305 & \best{0.190} & \second{357.20} \\
Ours & \best{24.156} & \second{0.767} & \best{0.259} & \best{0.157} & \best{0.571} & \second{0.214} & \best{353.28} \\
\bottomrule
\end{tabular}%
}
\end{center}
\caption{Quantitative comparison with state-of-the-art methods under heavier test degradation settings. \best{Best}, \second{second-best}, and \third{third-best} results are highlighted.}
\end{table*}